\documentclass[letterpaper]{article} % DO NOT CHANGE THIS
\usepackage{aaai2026}  % DO NOT CHANGE THIS
\usepackage{times}  % DO NOT CHANGE THIS
\usepackage{helvet}  % DO NOT CHANGE THIS
\usepackage{courier}  % DO NOT CHANGE THIS
\usepackage[hyphens]{url}  % DO NOT CHANGE THIS
\usepackage{graphicx} % DO NOT CHANGE THIS
\usepackage{natbib}  % DO NOT CHANGE THIS AND DO NOT ADD ANY OPTIONS TO IT
\usepackage{caption} % DO NOT CHANGE THIS AND DO NOT ADD ANY OPTIONS TO IT
\usepackage{algorithm}

\usepackage{newfloat}
\usepackage{listings}
\DeclareCaptionStyle{ruled}{labelfont=normalfont,labelsep=colon,strut=off} % DO NOT CHANGE THIS
\floatstyle{ruled}
\newfloat{listing}{tb}{lst}{}
\floatname{listing}{Listing}

\usepackage{amssymb}
\usepackage{amsmath}
\DeclareMathOperator*{\argmax}{arg\,max}
\DeclareMathOperator*{\argmin}{arg\,min}

\renewcommand{\th}{\mathbf{\theta}}
\usepackage{enumitem}
\usepackage{tcolorbox}
\usepackage{titlesec}
\tcbuselibrary{breakable}
\newtcolorbox{promptbox}[1][]{%
  colback=blue!10,
  colframe=blue!50!black,
  boxrule=0.5pt,
  arc=2mm,
  left=2mm, right=2mm,
  top=1mm, bottom=1mm,
  colbacktitle=blue!30!white,
  coltitle=black,
  fonttitle=\bfseries,
  toptitle=1mm,      % space above the title text
  bottomtitle=1mm, 
  title=#1,
}

\title{OptiHive: Ensemble Selection for LLM-Based Optimization via Statistical Modeling}
\author{
    Maxime Bouscary\textsuperscript{\rm 1} \quad Saurabh Amin\textsuperscript{\rm 1}
}
\affiliations{
    \textsuperscript{\rm 1}Massachusetts Institute of Technology, Cambridge, MA, USA \\
    \texttt{\{mbscry,amins\}@mit.edu}
}

\usepackage{bibentry}
\renewcommand{\P}[1]{\mathbb{P}\left(#1\right)}
\usepackage{booktabs}
\usepackage{algorithmic}
\usepackage{subcaption}

\begin{document}

\maketitle

\begin{abstract}
LLM-based solvers have emerged as a promising means of automating problem modeling and solving. However, they remain unreliable and often depend on iterative repair loops that result in significant latency. We introduce OptiHive, a framework that enhances any solver-generation pipeline to produce higher-quality solvers from natural-language descriptions of optimization problems. OptiHive uses a single batched generation to produce diverse components (solvers, problem instances, and validation tests) and filters out erroneous components to ensure fully interpretable outputs. Accounting for the imperfection of the generated components, we employ a statistical model to infer their true performance, enabling principled uncertainty quantification and solver selection. On tasks ranging from traditional optimization problems to challenging variants of the Multi-Depot Vehicle Routing Problem, OptiHive significantly outperforms baselines, increasing the optimality rate from 5\% to 92\% on the most complex problems.
\end{abstract}

\begin{links}
    \link{Code}{https://github.com/mbscry/OptiHive}
    % \link{Datasets}{https://aaai.org/example/datasets}
    \link{Extended version}{https://arxiv.org/abs/2508.02503}
\end{links}

\section{Introduction}

Large language models (LLMs) have demonstrated remarkable abilities in reasoning, code generation, and problem solving across a wide range of domains \cite{brown2020language, achiam2023gpt, chen2021evaluating, li2022competition}, making them increasingly relevant tools for tackling complex computational tasks. However, their application to complex optimization tasks remains hindered by unreliable code generation and self-evaluation. In practice, generated solvers often exhibit two distinct failure modes: hard computational errors (invalid syntax or runtime failure) that render code outright unusable, and soft quality deficiencies (incorrect algorithms, suboptimal solutions) that cannot be detected deterministically.

Traditional LLM optimization pipelines rely on iterative generate-evaluate-repair loops or fine-tuned models with integrated correction routines. These methods can address syntactic failures, but struggle to quantify the quality of solutions due to overconfidence in self-assessment. Moreover, the best-performing pipelines tend to involve long and often unpredictable latency, that is, the time between receiving a problem description and returning a final solver. Test-based methods, which prompt an LLM to generate input-output pairs or simple verification functions using hard-coded values, can improve assessment in simple settings, but often fail to produce valid test cases for complex optimization problems, where ground-truth outputs cannot be obtained without solving the problem itself.

In this work, we present OptiHive, a two-stage framework that enhances any solver-generation pipeline by decoupling interpretability constraints from quality uncertainty, enabling both deterministic elimination of unusable components and statistical inference over the remaining ones. In the first stage, we simultaneously generate multiple candidate solvers, problem instances, and validity tests. Interpretability, defined as the ability of a solver-instance-test triple to compile, execute, and return a syntactically valid result, is treated as a hard feasibility requirement. We formulate an integer linear program (ILP) to eliminate flawed components (solvers, instances, and tests) and retain interpretable outputs only. The second stage then applies a latent-class model over the filtered elements, jointly estimating latent ground truth variables (instance feasibility and solution validity) as well as the error rates of solvers and tests. The framework is agnostic to the optimization class and can be applied to linear, convex, and nonconvex problems, including continuous and mixed-integer formulations.

By quantifying the performance and trustworthiness of each solver, OptiHive provides principled uncertainty estimation and enables informed solver selection. Unlike agent-based methods and multi-stage prompt pipelines that incrementally refine a single solution, our statistical ensemble method departs from the prevailing ``generate-then-fix'' paradigm: it generates in parallel a set of candidate solvers from a given solver-generation pipeline, ranks the candidates by their inferred true quality, and selects the most promising one. OptiHive serves as a wrapper around existing solver-generation pipelines, substantially improving performance through rigorous statistical inference rather than self-critique. With minimal computational burden from filtering, inference, and selection, OptiHive adds negligible latency to the pipeline it wraps, and enables high performance with minimal overhead when paired with lightweight generation procedures such as single-shot LLM inference.

The framework relies on the solver generation step to produce at least one correct solver. While deterministic sampling (e.g., temperature 0) may fail to yield any valid candidate, higher temperatures can introduce the diversity needed to uncover correct solvers. As a result, OptiHive can recover valid solutions even when deterministic generation fails.

In summary, our work makes the following contributions:
\begin{enumerate}
    \item \textbf{Minimal and consistent latency through single batched inference and parallelization}. Our framework generates, in a single parallel step, independent candidate solvers, problem instances, and validity tests, eliminating iterative self-correction loops. Combined with fully parallelizable cross-evaluation of solutions and tests, this design yields high-quality solvers with minimal latency.
    \item \textbf{Reliable solver selection via statistical inference}. Previous work assumes that LLM self-evaluation provides trustworthy quality assessment, but recent research suggests that this assumption is fundamentally flawed due to LLMs' poor self-critique abilities and systematic biases \cite{stechly2024self, chen2024magicore, moon2025don}.
    Our framework addresses this by modeling all components (solvers, instances, and tests) as inherently noisy and employing rigorous statistical methods to estimate their true performance.
    \item \textbf{Numerical experiments on two classes of complex optimization problems}. We demonstrate that our framework reliably identifies high-quality solvers and significantly outperforms baselines on complex variants of the Multi-Depot Vehicle Routing Problem and the Weighted Set Cover Problem. 
\end{enumerate}

\section{Related Work}

Our work lies at the intersection of two streams of research within this field: LLMs for optimization problems and LLM-generated test functions.

\textbf{LLMs for Optimization Problems}

Chain-of-Thought (CoT) prompting \cite{wei2022chain} demonstrated that eliciting intermediate reasoning steps significantly improves performance on complex reasoning tasks in large-scale LLMs. Building on this, Composition of Experts (CoE) \cite{CoE} introduced an agent-based orchestrator that routes inputs to specialized LLM experts, improving scalability and customization in optimization pipelines. OptiMUS \cite{OptiMUS} and OptimAI \cite{OptimAI} present end-to-end frameworks that automatically translate natural language problem descriptions into mathematical models, validating the feasibility of large-scale optimization modeling. Both approaches incorporate self-correction loops, prompting the LLM to iteratively evaluate and revise its output until it passes a specified evaluation or reaches a maximum number of iterations. While this iterative refinement improves the correctness of generated models, it introduces significant latency and can be unreliable, particularly when the model repeatedly makes similar errors. Aligned with improvement loops, Optimization by PROmpting (OPRO)~\cite{yang2023large} proposes new candidates based on all previously generated solutions with their objective values, while Self-Guiding Exploration (SGE) \cite{iklassov2024self} employs a thought-based framework that breaks each problem into simpler subtasks and then iteratively refines each ``thought'' through a fixed number of LLM calls. Drawing on evolutionary algorithms, Hercules \cite{wu2025efficient} employs an outer loop that generates, evaluates, and selects heuristic candidates across successive generations.  Parallel to these methods, LLMOPT \cite{LLMOPT} and LLaMoCo \cite{Llamoco} enhance the problem formulation and solver code generation via instruction-tuning. In contrast, our framework adopts a linear approach that eliminates self-correction loops by producing a set of independent candidate solvers, filtering out solvers returning non-interpretable results, and selecting the highest potential solver based on the parameters estimated by the latent-class model. This design achieves both low latency and high performance, overcoming a key bottleneck of existing LLM-driven optimization pipelines.

\textbf{LLM-Generated Test Functions}

Recent studies indicate that LLMs are poor and biased in self-critique \cite{huang2023large, stechly2024self, chen2024magicore, moon2025don}. As a result, the automatic generation of external test functions, rather than relying on the model's self-evaluation, has gained traction. Most works focus on unit test generation \cite{chen2022codet, jain2024livecodebench, prasad2025learning, chen2024chatunitest, lin2025learning}, where LLMs are prompted to write valid input-output pairs. These approaches significantly reduce manual testing effort and can achieve high code coverage, but are not amenable to complex optimization problems where the reasoning capabilities of the LLM are insufficient to produce valid input-output pairs. Only a handful of efforts have explored generating complete test functions \cite{yuan2023no, schafer2023empirical, rao2023cat}, but these remain restricted to simple assert-based checks over hard-coded inputs. In contrast, we generate reusable test functions that take input-output pairs and verify problem-specific invariants, such as constraint feasibility and objective value consistency. The decoupling of tests from specific inputs (i.e. problem instances) makes our framework thrifty, as each test can be reused to validate diverse solver-instance pairs. 
% This makes OptiHive a framework-agnostic wrapper that extracts the best-performing solver from any code synthesis approach.

\section{Methodology}

\begin{figure*}[ht]
\centering
\includegraphics[width=0.95\textwidth]{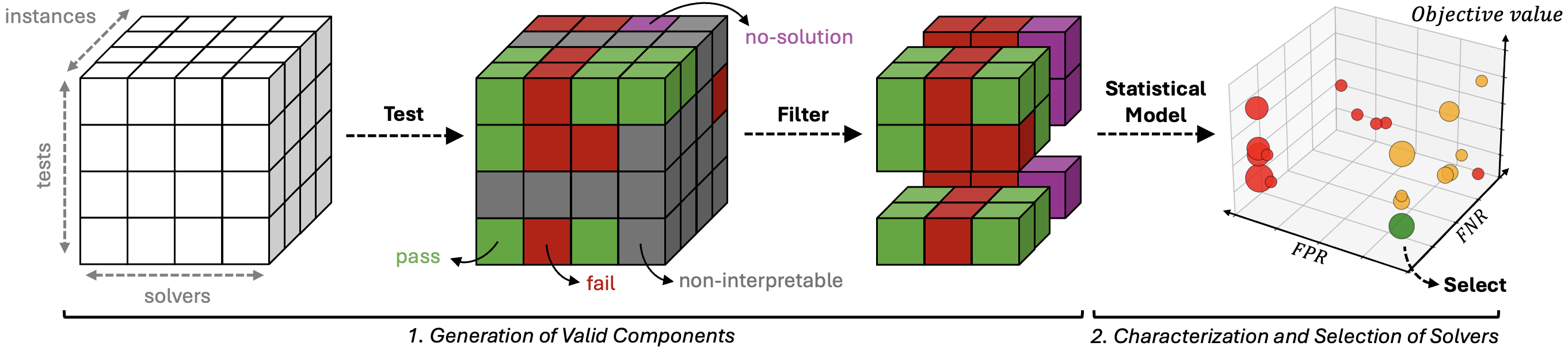} % Reduce the figure size so that it is slightly narrower than the column.
\caption{OptiHive produces optimization solvers through a two-stage process. In the first stage, it produces candidate solvers, problem instances, and tests, then filters out components to retain only fully interpretable solver-instance-test triples (represented as cubes). In the second stage, it applies latent class analysis to estimate the performance of each solver and selects the most promising candidate. Red, yellow, and green bubbles denote solvers that return non-interpretable or infeasible, feasible but suboptimal, and optimal solutions, respectively.}
\label{fig:framework}
\end{figure*}

Figure \ref{fig:framework} illustrates the two stages of OptiHive. In the first stage (generation of valid components), we efficiently produce executable components, resulting in a set of interpretable outputs. Specifically, we perform a single batch generation to obtain a pool of candidate solvers, problem instances, and validity tests. We compile and execute every solver-instance pair, and execute all tests on solver–instance pairs producing a solution. Then, we solve an ILP to remove a subset of solvers, instances, and tests, so that only interpretable outputs are retained. In the second stage (solver characterization and selection), we fit a latent-class model \cite{dawid1979maximum} on these interpretable outputs to jointly infer the feasibility of instances and each solver's solution validity, and the type I/II error rates of both solvers and tests. Finally, we rank solvers with a scalarized objective that integrates their estimated error rates and expected objective values, and select the top-ranked candidate. The entire framework is given in Algorithm \ref{alg:optihive}. We now describe each stage in detail, beginning with the generation of components.

\begin{algorithm}[ht]
\caption{OptiHive}
\label{alg:optihive}
\begin{algorithmic}[1]
\REQUIRE Sample sizes $N_S, N_I, N_T$
\STATE \textit{// Generation step}
\STATE Perform single batch generation to obtain:
\STATE \quad solvers $\bar{S}$ of size $N_S$
\STATE \quad instances $\bar{I}$ of size $N_I$
\STATE \quad tests $\bar{T}$ of size $N_T$
\STATE \textit{// Testing step}
\FORALL{$(s,i)\in \bar{S}\times \bar{I}$} 
    \STATE Execute solver $s$ on instance $i$ to obtain $r_{s,i}$
    \IF{$r_{s,i} = 1$}
        \FORALL{$t \in \bar{T}$}
            \STATE Execute test $t$ on solution of $(s,i)$ to obtain $r_{s,i,t}$
        \ENDFOR
    \ENDIF
\ENDFOR
\STATE \textit{// Filtering step}
\STATE Solve \eqref{eq:ILP} to obtain $x^\star$, $y^\star$, $z^\star$
\STATE $S \gets \{s \in \bar{S}: x^\star_s = 1\}$
\STATE $I 
\gets \{i \in \bar{I}: y^\star_i = 1\}$
\STATE $T \gets \{t \in \bar{T}: z^\star_t = 1\}$
\STATE $\mathbf{R} 
\gets \{r_{s,i,t}:(s,i,t) \in S\times I \times T\}$
\STATE \textit{// Characterization step}
\STATE Initialize $\theta_0 = (\lambda, \{\alpha_s,\beta_s,\gamma_s\}_{s \in S}, a_0, b_0, a_1, b_1)$
\REPEAT
    \STATE Compute $\theta_{k+1}$ from \eqref{eq:EM_update} with $\theta_k$ and $\mathbf{R}$
\UNTIL{convergence to $\theta^\star$ or iteration limit}
\STATE \textit{// Selection step}
\STATE $s^\star \gets \argmin_{s \in S} g(\theta^\star, s)$
\RETURN $s^\star$
\end{algorithmic}
\end{algorithm}

\subsection{Generation of Valid Components}

\subsubsection{Batched Components Generation.}

Given a problem description with input and output format specifications, we invoke an LLM to generate three sets:
\begin{enumerate}
    \item Candidate Solvers $\bar{S}$. Each solver $s \in \bar{S}$ is a function taking as input a problem instance $i$ and returning an optimization report with \textsc{infeasible} status if the instance admits no feasible solution, or the best solution found within the time limit at its corresponding objective value with either \textsc{optimal} or \textsc{time\_limit} status.
    \item Problem Instances $\bar{I}$. The diversity of instances in $\bar{I}$ is encouraged by providing different random seeds in each prompt, and slightly varying the phrasing of the prompts, explicitly requesting feasible, infeasible, or random instances.
    \item Validity Tests $\bar{T}$. Each test $t \in \bar{T}$ is a function taking as input an instance-solution pair $(i, x)$ and returning \textsc{True} if and only if the solution $x$ is feasible for $i$ and its true objective value matches the reported objective value.
\end{enumerate}

All prompts are provided in Appendix A of the extended version of this paper. Importantly, every component in $\bar{S}$, $\bar{I}$, and $\bar{T}$ is independent of other components, so that all components can be generated simultaneously. The set $\bar{S}$ can be obtained using a more refined generation pipeline. In any case, no additional calls to the LLM are required during subsequent steps, which is key to the low latency of our framework.

\subsubsection{Testing.}

Note that all LLM-generated components are subject to being syntactically or semantically invalid. For each tuple $(s,i) \in \bar{S} \times \bar{I}$, we attempt to obtain an optimization report by running $s$ on $i$. We say that the tuple $(s,i)$ is \emph{interpretable} if $s$ and $i$ compile, running $s$ on $i$ does not raise an error during execution, and the report of $(s,i)$ contains a \emph{status} field with an interpretable value. For every test $t$, we say that the triple $(s,i,t)$ is \emph{interpretable} when test $t$ compiles and either:
\begin{enumerate}
    \item $(s,i)$ is interpretable with a report without solution.
    \item $(s,i)$ is interpretable with a report containing a solution, running test $t$ on the solution of $(s,i)$ does not raise error during execution, and returns a boolean.
\end{enumerate}

\subsubsection{Filtering.}

Instead of integrating a costly (and often unreliable) self-correction loop for each component to ensure compilability, executability, and evaluability, we filter out a subset of the components to retain only interpretable triples $(s,i,t)$.

To retain a maximum number of components while filtering out all the non-interpretable triples, we formulate the following ILP:
\begin{alignat}{3}\label{eq:ILP}
    \max_{x, y, z} \quad & \sum_{s \in \bar{S}} x_s + \sum_{i \in \bar{I}} y_i + \sum_{t \in \bar{T}} z_t\\
    \text{s.t.} \quad & x_s + y_i + z_t \leq 2, && \forall (s,i,t) \in \mathcal{U}\nonumber\\
    &x_s \in \{0,1\}, && \forall s \in \bar{S}\nonumber\\
    &y_i \in \{0,1\}, && \forall i \in \bar{I}\nonumber\\
    &z_t \in \{0,1\}, && \forall t \in \bar{T}\nonumber
\end{alignat}
where $x_s$ (resp. $y_i$ and $z_t$ ) are binary variables equal to 1 if and only if solver $s$ (resp. instance $i$ and test $t$) are kept, and $\mathcal{U} = \{(s,i,t) \in \bar{S} \times \bar{I} \times \bar{T}: \text{ $(s,i,t)$ is not interpretable}\}$. After solving \eqref{eq:ILP}, we obtain optimal selections $x^\star$, $y^\star$, and  $z^\star$ and define:
\begin{align*}
    S &\triangleq \{s \in \bar{S}: x^\star_s = 1\}\\
    I &\triangleq \{i \in \bar{I}: y^\star_i = 1\}\\
    T &\triangleq \{t \in \bar{T}: z^\star_t = 1\}
\end{align*}
By construction, every triple in $S \times I \times T$ is interpretable. This ILP is tractable even with hundreds of solvers, instances, and tests, as the number of variables grows linearly with the number of components, and $\mathcal{U}$ exhibits a highly structured pattern since failure of a solver, instance, or test often induces multiple non-interpretable triples involving that component.

\subsection{Solver Characterization and Selection}

\subsubsection{Characterization.}

Although every triple $(s,i,t)$ is now interpretable, we only observe solver reports and, when a report contains a solution, whether that solution passes a suite of imperfect tests. Notably, the true feasibility of instances and the validity of reported solutions are unknown, as both solvers and tests are generated by an LLM and thus cannot be assumed to be perfectly trustworthy. By treating these unobserved variables as latent, we use a latent-class model to jointly estimate the membership of instances and solutions, the accuracy of solvers (type I/II error rates, validity rate of solutions on feasible instances), and the accuracy of tests (type I/II error rates). 
Specifically, we observe for all $(s,i) \in S \times I$:
\begin{align*}
    r_{s,i} &= \left\{\begin{tabular}{cl}
        1 & \text{if solver $s$ reports a solution}\\
        0 & \text{otherwise.}
    \end{tabular}
    \right.
\end{align*}
and for all $(s,i,t) \in S \times I \times T$ such that $r_{s,i} = 1$:
\begin{align*}
    r_{s,i,t} &= \left\{\begin{tabular}{cl}
        1 & \text{ if the solution of $(s,i)$ passes test $t$}\\
        0 & \text{otherwise.}
    \end{tabular}
    \right.
\end{align*}

For all $i \in I$, we define the latent variable:
\begin{align*}
    f_{i} &= \left\{\begin{tabular}{cl}
        1 & \text{if instance $i$ admits a feasible solution}\\
        0 & \text{otherwise.}
    \end{tabular}
    \right.
\end{align*}
and for all $(s,i) \in S \times I$ such that $r_{s,i} = 1$:
\begin{align*}
    f_{s,i} &= \left\{\begin{tabular}{cl}
        1 & \text{if $r_{s,i}=1$ and the solution of $(s,i)$ is feasible}\\
        0 & \text{otherwise.}
    \end{tabular}
    \right.
\end{align*}

We then make the following assumptions on the distribution of these variables:
\begin{enumerate}
    \item Instances. For all $i \in I$:
    $$\P{f_i = 1} = \lambda$$
    with $\lambda$ being the probability that a randomly drawn instance is feasible.
    \item Solvers. For all $(s,i) \in S \times I$:
    \begin{align*}
        &\P{r_{s,i} = 1 \mid f_i = 0} = \alpha_s\\
        &\P{r_{s,i} = 1 \mid f_i = 1} = 1-\beta_s\\
        &\P{f_{s,i} = 1 \mid r_{s,i} = 1, f_i = 1} = \gamma_s
    \end{align*}
    where $\alpha_s$, $\beta_s$, and $\gamma_s$ are false positive rate, false negative rate, and rate of feasible solutions among reports on positive instances, respectively, for solver $s$.
    \item Tests. For each solver-instance pair with $r_{s,i} = 1$, we aggregate the test outcomes as $C_{s,i} = \sum_{t \in T} r_{s,i,t}$. Test outcomes are not independent since some solutions are inherently harder to evaluate than others. However, conditioned on the latent variable $f_{s,i}$, the individual test results are assumed to be exchangeable. To capture this, we model $C_{s,i}$ with Beta-Binomial distributions as:
    \begin{align*}
        &C_{s,i} \mid f_{s,i} = 0 \sim \text{BetaBinomial}\left(|T|, a_0, b_0\right)\\
        &C_{s,i} \mid f_{s,i} = 1 \sim \text{BetaBinomial}\left(|T|, a_1, b_1\right)
    \end{align*}
\end{enumerate}
To break the symmetry, we leverage the fact that tests rarely reject feasible solutions and use a strong prior $\text{Beta}(\bar{\alpha},\bar{\beta})$ with $\bar{\alpha} \gg \bar{\beta}$ on the tests' true-positive rate $\frac{a_1}{a_1 + b_1}$, which strongly favors values close to 1.

Let $\theta = \left(\lambda, \{\alpha_s, \beta_s, \gamma_s\}_{s \in S}, a_0, b_0, a_1, b_1\right)$ denote the set of all parameters, and $\mathbf{R}$ (resp. $\mathbf{F}$) be the set of observed (resp. latent) variables. We use the expectation–maximization (EM) algorithm to find a set of parameters $\theta^\star$ locally maximizing the observed data likelihood function by iterating over the following update
\begin{equation}\label{eq:EM_update}
    \theta_{k+1} = \argmax_\theta \mathbb{E}_{\mathbf{F} \sim \P{\cdot \mid \mathbf{R}, \theta_{k}}} \left[\ln \P{\mathbf{R}, \mathbf{F} \mid \theta}\right]
\end{equation}
until convergence. The distribution of the latent variables $\{f_i\}_{i \in I}$, $\{f_{s,i}\}_{(s,i) \in S \times I}$ can then be estimated from $\theta^\star$. See Appendix B in the extended version for details on update \eqref{eq:EM_update}.

\subsubsection{Selection.}

For all reports containing a solution, let $z_{s,i}$ be the objective value reported by solver $s$ on instance $i$, and
$$Z_s = \mathbb{E}_{i \sim \P{\cdot \mid r_{s,i} = 1, f_{s,i} = 1}}\left[ z_{s,i} \right]$$ be the conditional expected objective over feasible solutions reported by solver $s$. 

Solver selection requires balancing multiple, often conflicting objectives. While some solvers may reliably detect infeasible instances, others may excel at returning high-quality solutions on feasible ones, yet fail to distinguish infeasibility. Relying solely on expected objective value can favor solvers that over-report feasibility or silently fail on hard instances.

% The solver selection can be formulated as a multi-objective unconstrained optimization problem using the set of parameters $\theta^\star$:
% \begin{equation}\label{eq:multiobjective_problem}
%     \min_{s \in S} \left(\alpha_s, \beta_s, 1-\gamma_s, Z_s\right).
% \end{equation} Since there does not typically exist a solver that minimizes all objectives simultaneously, the final solver is selected from the Pareto front induced by \eqref{eq:multiobjective_problem} depending on the objective weights or priorities. In the absence of an a priori method, we propose to select the final solver by scalarizing \eqref{eq:multiobjective_problem} as:

To account for this trade-off, we define a scalarized objective function that summarizes overall solver quality in a single score:
\begin{multline}\label{eq:scalarized}
  g(\theta^\star, s) \triangleq \lambda (1-\beta_s)\gamma_s Z_s + \lambda \beta_s P_\text{miss} \\[\jot]
  + \left((1-\lambda)\alpha_s + \lambda (1-\beta_s)(1-\gamma_s)\right) P_\text{fail}
\end{multline}
Specifically, $g(\theta^\star, s)$ denotes the expected objective value of solver $s$ under penalties $P_\text{miss}$ when no solution is found for a feasible instance, and $P_\text{fail}$ when an infeasible solution is reported, respectively. By default, $P_{\text{miss}}$ and $P_{\text{fail}}$ can be set to $10 \cdot Z_\text{max}$, with $Z_\text{max}$ defined as the maximum absolute objective value reported across all solver-instance pairs. This ensures that both under-reporting and over-reporting solvers are severely penalized. This scalarized objective provides a basis for ranking solvers, from which the final solver is selected as $s^\star = \argmin_{s\in S} g(\theta^\star, s)$.

\section{Experimental Results}

Previous benchmark datasets for optimization from natural language, such as NLP4LP \cite{ahmaditeshnizi2023optimus} and ComplexOR \cite{CoE}, include problems that recent LLMs can now solve reliably, but where perfect scores remain difficult to achieve due to ambiguous or underspecified problem statements. For example, instances may describe integer-valued quantities while asking for LP formulations, leaving the model to infer the intended discretization. As a result, they do not allow for a reliable evaluation of solver performance, since failures may stem from prompt ambiguity rather than solver quality.
To address this, we designed variants of the Multi-Depot Vehicle Routing Problem (MDVRP) and the Weighted Set Cover Problem (WSCP) that introduce increasing levels of complexity, enabling meaningful performance evaluation.

\subsection{Experimental Setup}

We compare the performance of solvers returned by OptiHive against those produced directly by the same LLM used in the solver generation step, in order to isolate and highlight the marginal gains provided by OptiHive's selection mechanism. Rather than comparing against existing LLM-based optimization pipelines, our focus is on demonstrating how OptiHive can enhance the performance of any such pipeline by selecting high-quality solvers from a pool of candidates.

For each problem, we sample random tractable instances and compute ground-truth objectives using a reference solver. A candidate solver is \emph{feasible} if all its solutions satisfy the variant-specific constraints, and \emph{optimal} if, in addition, the reported objectives match the ground truth within a specified tolerance.

We generate sets of 100 solvers, 100 instances, and 100 tests per problem, then pre-compute the outputs of all solver-instance pairs and all solver-instance-test triples corresponding to interpretable pairs. We evaluate the performance of OptiHive by sampling, with replacement, the specified number of components, and proceed with filtering, characterizing, and selecting a solver. We repeat this process 10,000 times with random seeds for every run to obtain a reliable estimate of the performance.

We use OpenAI models (\texttt{gpt-4.1-nano}, \texttt{gpt-4.1-mini}, and \texttt{o3}), with the temperature fixed at the default value of 0.7. The specific model used is indicated for each problem. Runs are parallelized on an AMD EPYC 9734 processor. The number of expectation-maximization loops is limited to 100 iterations, and all individual runs complete in less than 1 second. 
This runtime is negligible compared to the time required for LLM inference or the overhead incurred by typical code generation pipelines.

\subsection{Multi-Depot Vehicle Routing Problems}

We study two variants of MDVRP \cite{toth2002vehicle}:
\begin{itemize}
    \item Distance-Constrained Multi-Depot Vehicle Routing Problem (DCMDVRP)
    \item Multi-Depot Vehicle Routing Problem with Obstacles (MDVRP+OBS)
\end{itemize}
In the DCMDVRP, each vehicle's total travel distance is constrained by a fixed upper bound. While this variant is less common than the standard MDVRP, it can typically be addressed by adding a single distance constraint per vehicle. Aside from rare formulation flaws, such as improperly allowing vehicles to reset their cumulated distance when visiting the depot of another vehicle, the formulation of this variant is conceptually very similar to MDVRP.
In contrast, the MDVRP+OBS introduces significant geometric complexity: the 2D space contains line-segment obstacles that vehicles cannot cross, fundamentally altering the feasible routing space. Solvers that correctly address this problem typically follow a structured yet non-trivial pipeline: (i) augmenting the node set with obstacle endpoints, (ii) computing shortest paths via a visibility graph that excludes edges intersecting obstacles, (iii) solving a standard MDVRP using the resulting distance matrix, and (iv) mapping the ILP solution back to explicit routes. This process yields long, intricate code prone to oversimplification (e.g., by removing invalid edges instead of properly computing obstacle-aware shortest paths), which often leads to crashing, invalid, and suboptimal solutions.

We use \texttt{gpt-4.1-mini} and \texttt{o3} to generate the components of the DCMDVRP and MDVRP+OBS, since \texttt{o3} consistently produces optimal solvers on DCMDVRP and \texttt{gpt-4.1-mini} never produced optimal solutions on MDVRP+OBS.

\begin{figure*}[t]
  \centering
  \begin{subfigure}[c]{0.44\textwidth}
    \centering
    \subcaptionbox{DCMDVRP: Feasibility and optimality\label{fig:dcmdvrp}}{
      \includegraphics[width=0.48\textwidth]{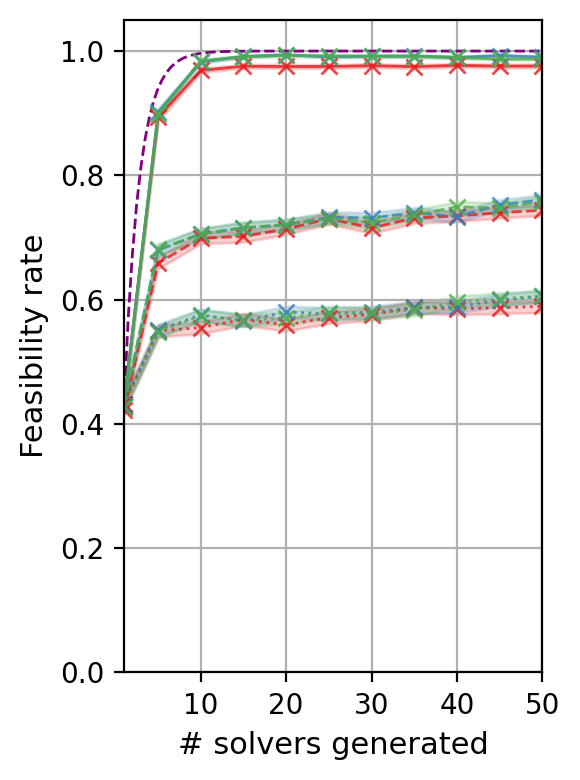}
      \includegraphics[width=0.48\textwidth]{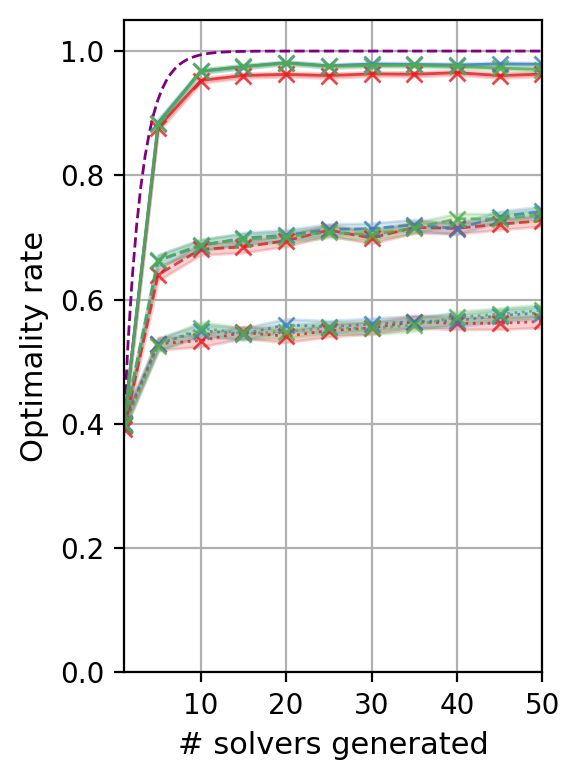}
    }
  \end{subfigure}
  \hfill
  \begin{subfigure}[c]{0.1\textwidth}
    \centering
    \includegraphics[width=\textwidth]{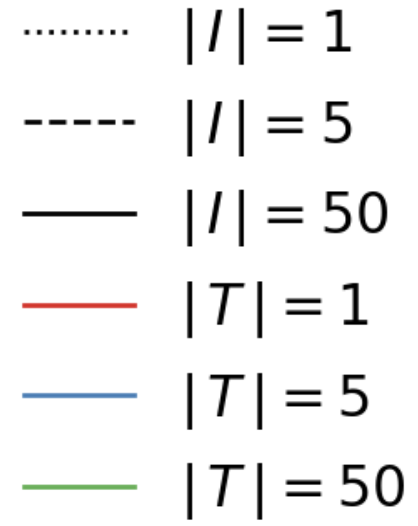}
  \end{subfigure}
  \hfill
  \begin{subfigure}[c]{0.44\textwidth}
    \centering
    \subcaptionbox{MDVRP+OBS: Feasibility and optimality\label{fig:mdvrpobs}}{
      \includegraphics[width=0.48\textwidth]{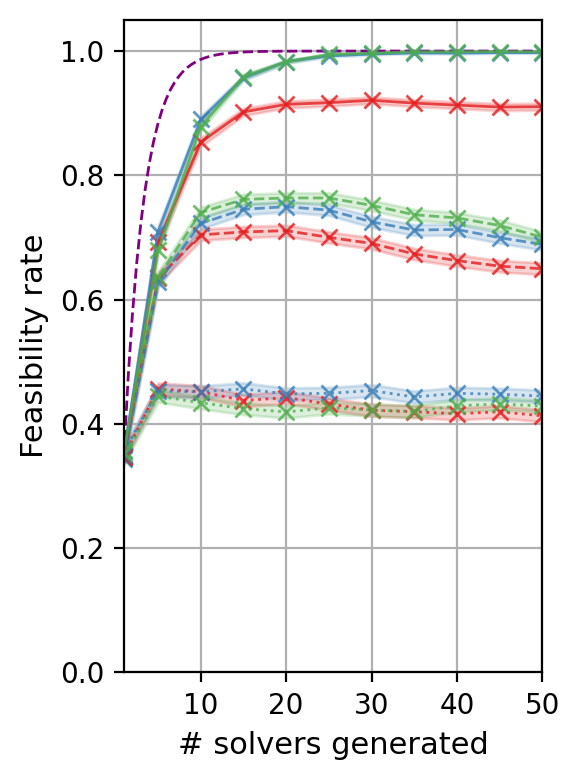}
      \includegraphics[width=0.48\textwidth]{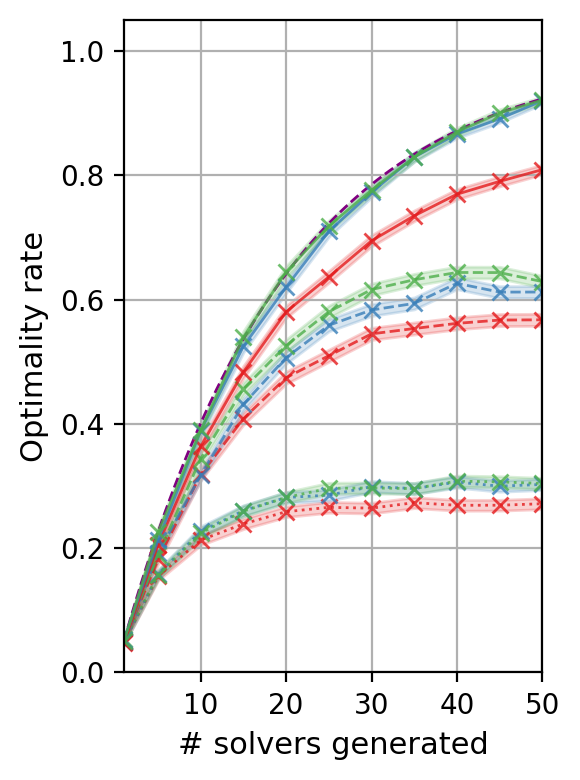}
    }
  \end{subfigure}
  \caption{Feasibility and optimality rates on DCMDVRP and MDVRP+OBS across varying numbers of generated solvers, instances, and tests. The purple curve shows the rate under perfect selection i.e. the probability that at least one of the generated solvers is optimal.}
  \label{fig:results_vrp}
\end{figure*}

Figure~\ref{fig:results_vrp} shows the optimality rate of the solver selected via OptiHive as the number of candidate solvers increases, for varying numbers of generated instances and tests.

Quantitatively, OptiHive significantly improves both feasibility and optimality over the LLM baseline generating a single solver. When using 50 components of each type, feasibility rises from 43\% to 98.74\% and optimality from 40\% to 97.03\% on DCMDVRP. On the considerably harder MDVRP+OBS, feasibility improves from 35\% to 99.88\% and optimality from only 5\% to 92.1\%.

We observe that the number of generated instances is critical. Diverse instances create a richer signal for the latent-class model and are necessary to differentiate optimal solvers from near-optimal or subtly flawed solvers that are correct on most instances but fail on corner or edge cases. Sophisticated heuristics typically require sufficiently complex and specific instances to be distinguished from optimal solvers.

The number of solvers is, as expected, another key ingredient to the performance of our framework. On simpler problems such as DCMDVRP, where only a handful of generated solvers already include an optimal one with high probability, performance plateaus quickly once the solver pool reaches a modest size (\ref{fig:dcmdvrp}). In contrast, for more challenging problems, the number of solvers is essential: sampling many more solvers markedly increases the probability of including at least one optimal candidate, and thus has a much greater impact on overall performance (\ref{fig:mdvrpobs}).

On easy to moderately complex problems, most candidate solvers cluster around the true optimal solution and form a consensus, while the flawed ones are scattered and behave like noise. This structure makes it easier for the latent model to identify the optimal solvers (\ref{fig:dcmdvrp}). In contrast, the opposite can occur on hard problems: many solvers may concentrate around incorrect solutions, with the few optimal ones appearing as outliers. In such cases, adding more solvers can degrade performance, especially if the set of instances is too small or not diverse enough to retrieve optimal solvers from the dominant and incorrect consensus (\ref{fig:mdvrpobs}).

Finally, the number of tests yields the smallest improvement in performance. Evaluating solution correctness is generally easier than generating a solution, and tests tend to be more reliable. As a result, the performance saturates quickly as more tests are added. In our experiments, we observe marginal improvement beyond five generated tests: once all failure modes are covered, additional tests add little value. While test diversity still matters to avoid blind spots and guard against rare flawed tests, we note that the cost–benefit curve for adding tests is substantially flatter than for increasing the number of solvers or instances in this experiment.

\subsection{Weighted Set Cover Problem}

The WSCP \cite{chvatal1979greedy, caprara2000algorithms} can be illustrated through a practical scenario involving emitters and clients. Each emitter is characterized by a location, radius, and activation cost. An emitter is said to cover a client if the client lies within its coverage range. The objective is to select a minimum-cost subset of emitters so every client is covered by at least one active emitter. We consider three variants:
\begin{itemize}
    \item $K$-robust WSCP: Coverage must survive any $K$ adversarial emitter failures
    \item Probabilistic WSCP: Emitter availability is uncertain and coverage is required in a probabilistic sense 
    \item Time-dependent WSCP: clients move over time, so activations must be scheduled to sustain coverage across a time horizon.
\end{itemize}

\begin{table*}[ht]
\centering
\begin{tabular}{l *{12}{c}}
\toprule
& \multicolumn{4}{c}{$K$-robust} & \multicolumn{4}{c}{Probabilistic} & \multicolumn{4}{c}{Time-dependent} \\
\cmidrule(lr){2-5} \cmidrule(lr){6-9} \cmidrule(lr){10-13}
& \multicolumn{2}{c}{Baseline} & \multicolumn{2}{c}{OptiHive}
& \multicolumn{2}{c}{Baseline} & \multicolumn{2}{c}{OptiHive}
& \multicolumn{2}{c}{Baseline} & \multicolumn{2}{c}{OptiHive} \\
\cmidrule(lr){2-3} \cmidrule(lr){4-5}
\cmidrule(lr){6-7} \cmidrule(lr){8-9}
\cmidrule(lr){10-11} \cmidrule(lr){12-13}
& Opt. & Feas. & Opt. & Feas.
& Opt. & Feas. & Opt. & Feas.
& Opt. & Feas. & Opt. & Feas. \\
\midrule
Reference & 98\% & 98\% & 100\% & 100\% & 73\% & 75\% & 100\% & 100\% & 3\% & 12\% & 64.1\% & 74.5\% \\
\texttt{nano} solvers & 42\% & 44\% & 83.1\% & 83.1\% & 33\% & 36\% & 89.9\% & 89.9\% & 0\% & 2\% & 0\% & 0.01\% \\
\texttt{nano} instances & 98\% & 98\% & 99.3\% & 99.3\% & 73\% & 75\% & 100\% & 100\% & 3\% & 12\% & 23.5\% & 37.7\% \\
\texttt{nano} tests & 98\% & 98\% & 100\% & 100\% & 73\% & 75\% & 100\% & 100\% &  3\% & 12\% & 19.7\% & 24.1\% \\
\bottomrule
\end{tabular}
\caption{Ablation study on the performance of OptiHive across variants of the Weighted Set Cover Problem.}
\label{tab:ablation}
\end{table*}

In the $K$-robust variant, the chosen emitters must ensure that every client stays covered even if an adversary deactivates any $K$ of them. While this may appear as a complex combinatorial requirement, it reduces to a simple condition: each client must be within range of at least $K+1$ selected emitters. This reformulation makes the problem equivalent to a standard WSCP with a minor modification to the coverage constraint.

The probabilistic variant introduces uncertain reliability of emitters. Here, each emitter $i$ fails independently with known probability $p_i$, and each client $j$ demands a minimum coverage probability $\pi_j$. Namely, if $x_i$ is a binary variable representing whether emitter $i$ is active, the solution must satisfy:
$$\mathbb{P}\left(\text{client }j\text{ covered}\right) = 1 - \prod_{i \in S_j\;:\;x_i=1} \left(1-p_i\right) \geq \pi_j, \quad \forall j$$
where $S_j$ is the set of emitters within range of client $j$. This makes the problem non-linear and intractable, but rearranging the terms and taking the log on both sides allows us to formulate equivalent but linear inequalities
$$\sum_{i \in S_j} x_i \log\left(1-p_i\right) \leq \log\left(1-\pi_j\right), \quad \forall j$$
which reduces the problem to an ILP.

The time-dependent variant models clients moving at constant speed along straight paths over a fixed horizon. Solving it typically involves: computing time intervals where each client is within range of an emitter (via a quadratic equation), identifying critical subintervals with changing coverage sets, solving static WSCP, and merging selected subintervals into contiguous activation schedules. These compounded complexities make the time-dependent variant the hardest to solve.

\textbf{Ablation study}. We perform an ablation study over the three component types: solvers, instances, and tests. The baseline method generates a single solver and returns it. For OptiHive, we sample 50 elements of each component type, run the EM algorithm, and return the solver that minimizes the scalarized objective in \eqref{eq:scalarized} with default penalties. We replace one component type at a time (solvers, instances, or tests) with generations from a smaller LLM to isolate the impact of each component's quality on overall performance. Table~\ref{tab:ablation} reports how this ablation affects both the baseline method and OptiHive. The reference setting corresponds to using \texttt{gpt-4.1-mini} to generate all three component types. In the other settings, we generate one component type with \texttt{gpt-4.1-nano} while keeping the other two generated by \texttt{gpt-4.1-mini}.

The results show that across all variants, OptiHive significantly improves both optimality and feasibility rates over the baseline generator, even when component quality degrades. This demonstrates that the latent-class model can recover signal from imperfect and noisy components. In the $K$-robust and probabilistic variants, reducing the quality of the set of instances or tests has little effect on performance. This supports our hypothesis that tests are \emph{generally} substantially easier to write correctly than the solvers themselves.

In contrast, the time-dependent variant suffers a substantial performance drop when the quality of either instances or tests is weakened. Generating a balanced set of feasible and infeasible instances is harder in this setting, and the weaker model produces an overwhelming number of infeasible instances instead of a more even split. Moreover, this variant exposes a scenario where validity tests are themselves difficult to produce. This challenges a core assumption of our method: that testing a solution is typically easier than generating one, allowing OptiHive to differentiate good solvers through test signals. Remarkably, even in this setting, tests generated by the small model still provide a noisy but meaningful signal that allows OptiHive to outperform the baseline. This highlights OptiHive's ability to extract value from an ensemble of entirely noisy components, a crucial feature for LLM-based optimization.

Lastly, OptiHive can successfully identify optimal solvers even when most candidates are infeasible. For the time-dependent WSCP, it raises the optimality rate from 3\% to 64.1\%. Note that with the choice $N_S=50$ solvers per run for this experiment, perfect selection would achieve a 78.2\% optimality rate. Conditioned on sampling at least one optimal solver in the pool of 50, we observe that OptiHive successfully selects an optimal solver 81.6\% of the time in the reference setting, and still obtains 30.9\% and 25.3\% success rates with degraded instances and degraded tests, respectively.

\section{Conclusion}

We introduced OptiHive, a two-stage framework that (i) generates candidate solvers, instances, and tests in parallel, evaluates, and filters out unusable components, and (ii) fits a latent-class model to infer the true quality of solvers, enabling calibrated solver selection. By efficiently extracting high-performing solvers from imperfect candidate pools, OptiHive wraps around any solver-generation pipeline to deliver higher-quality solvers with negligible added latency.

Empirical results show that OptiHive substantially improves performance on challenging optimization tasks. It raises the optimality rate on simple problems to near perfection and recovers underrepresented optimal solvers on the hardest problems, increasing optimality from 5\% to 92\%. Our ablation study further reveals that the quality of each component type -- solvers, instances, and tests -- plays a significant role in overall effectiveness. In particular, high-quality instances and tests provide a crucial signal for distinguishing optimal solvers. Still, even when components are produced by smaller models, OptiHive consistently delivers improved performance over baselines.

Future work includes exploring both heterogeneous solver sources and multi-stage generation strategies. These extensions could enable OptiHive to build richer solver ensembles and tackle even more challenging optimization problems.

\section{Acknowledgment}
We thank MIT Lincoln Laboratory for its support throughout this work. In particular, we are deeply grateful to Samuel A. Scheele and Andrew J. Weinert for their sustained collaboration and valuable feedback.

\bibliography{refs}

\clearpage 
\onecolumn 
\newcommand{\E}[1]{\mathbb{E}\left[#1\right]}
\setcounter{secnumdepth}{2}
\renewcommand\thesection{\Alph{section}}
\renewcommand{\thesubsection}{\thesection.\arabic{subsection}}
\setlength{\parskip}{6pt} 

\titleformat{\section}
  {\normalfont\Large\bfseries} % format of title text
  {\thesection}                % label (A, B, …)
  {1em}                        % space between label and title
  {}                           % before-code
  []

% \section{Maximum Triples MILP}

% The optimal selection of elements to remove to keep the maximum number of triples while removing all non-interpretable triples is achieved by solving the following MILP:
% \begin{alignat*}{3}
%     \max_{x, y, z, w} \quad & \sum_{(s,i,t) \in \bar{S} \times \bar{I} \times \bar{T}} w_{s,i,t}\\
%     \text{s.t.} \quad & x_s + y_i + z_t \leq 2, && \forall u(s,i,t) = 0\\
%     &w_{s,i,t} \leq x_s, && \forall (s,i,t) \in \bar{S} \times \bar{I} \times \bar{T}\\
%     &w_{s,i,t} \leq y_i, && \forall (s,i,t) \in \bar{S} \times \bar{I} \times \bar{T}\\
%     &w_{s,i,t} \leq z_t, && \forall (s,i,t) \in \bar{S} \times \bar{I} \times \bar{T}\\
%     &x_s \in [0,1], && \forall s \in \bar{S}\\
%     &y_i \in [0,1], && \forall i \in \bar{I}\\
%     &z_t \in [0,1], && \forall t \in \bar{T}\\
%     &w_{s,i,t} \in \{0,1\} && \forall (s,i,t) \in \bar{S} \times \bar{I} \times \bar{T}
% \end{alignat*}
% where $u(s,i,t) = 1$ if $(s,i,t)$ is interpretable, 0 otherwise. Note that the formulation involves $|\bar{S}| \cdot |\bar{I}| \cdot |\bar{T}|$ binary variables, which makes the formulation intractable when handling hundreds of solvers, instances, and tests.

\section{Prompt Templates}

\subsection{Solvers}

The following prompt is used to produce solvers given a \emph{problem description}, \emph{input template}, and \emph{output template}.

\begin{promptbox}[Solver Generation]
You are a code-generation agent expert in Python and Gurobi.\\

\textbf{[Problem Specifications]}\\
Here is the problem description: \{\texttt{problem\_description}\}

Here is the template for the problem input: \{\texttt{input\_template}\}

The output of the function must follow the template: \{\texttt{output\_template}\}\\

\textbf{[Instructions]}\\
Your task is to implement a function \texttt{solve} with a unique argument \texttt{data} as input and returning a solution to the problem.

Write the complete, executable, and well-indented code of the \texttt{solve} function, including necessary imports.

Status codes are: \texttt{OPTIMAL} for a proven best feasible solution, \texttt{INFEASIBLE} when no feasible solution is found.

Use a TimeLimit of 5 seconds for the optimization. Do not include example usage.
\end{promptbox}

\subsection{Instances}

We encourage instance diversity via two mechanisms. First, we rotate diversity directives among six options:
\begin{enumerate}
    \item If possible, the data should be an infeasible instance for the above problem.
    \item The data should be a clearly feasible instance for the above problem in that it admits a simple feasible solution.
    \item The data should result in optimal solutions to the above problem having tight constraints.
    \item The data should be randomized.
    \item The data should be randomized with hyperparameters that will make the instance likely feasible.
    \item The data should be randomized with hyperparameters that will make the instance likely infeasible.
\end{enumerate}
Second, we sample and provide a random seed to limit the similarity of numerical values across generated instances. To avoid lengthy LLM outputs, we ask the LLM to provide a function that returns an instance of the considered problem, rather than directly providing an instance.

We implement these guidelines through the prompt below.

\begin{promptbox}[Instance Generation]
You are a code‐generation agent expert in Python.\\

\textbf{[Problem Specifications]}\\
Consider the following problem: \{\texttt{problem\_description}\}\\
Here is the template for the problem input: \{\texttt{input\_template}\}\\

\textbf{[Instructions]}\\
Your task is to implement a function \texttt{generate\_input} with no argument and returning a input following the input template.

\{\texttt{diversity\_directives}\}

Write the complete, executable and well indented code of the \texttt{generate\_input} function, including necessary imports.

Use the following seed: \{\texttt{seed}\}
\end{promptbox}

\subsection{Tests}

\begin{promptbox}[Test Generation]
You are a code‐generation agent expert in Python.\\

\textbf{[Problem Specifications]}\\
Here is the problem description: \{\texttt{problem\_description}\}\\
Here is the input template: \{\texttt{input\_template}\}\\
Here is the solution template: \{\texttt{output\_template}\}\\

\textbf{[Instructions]}\\
For every concrete instance \texttt{data} that follows the input template, there is a corresponding \texttt{solution} object that follows the solution template.

Your task is to implement a function 
\texttt{test(data, solution) -> bool} 
that returns \texttt{True} if and only if all of the following hold:

\begin{enumerate}[nosep,left=1em]
  \item The solution is feasible (it satisfies every problem constraint).
  \item The reported objective value matches the cost you compute (within a small numerical tolerance).
  \item All solution fields are internally coherent.
\end{enumerate}

Write the complete, executable, and well‑indented Python code implementing the \texttt{test} function, including necessary imports.

Do not include example usage.
\end{promptbox}

\section{Expectation Maximization Model Specifications}

Observed variables:
\begin{align*}
    r_{s,i} &= \left\{\begin{tabular}{cl}
        1 & \text{if solver $s$ reports a solution}\\
        0 & \text{otherwise.}
    \end{tabular}
    \right., \quad \forall (s,i) \in S \times I\\
    r_{s,i,t} &= \left\{\begin{tabular}{cl}
        1 & \text{solution $(s,i)$ passes test $t$}\\
        0 & \text{otherwise.}
    \end{tabular}
    \right., \quad \forall (s,i,t) \in S \times I \times T \text{ s.t. } r_{s,i} = 1
\end{align*}
For all $(s,i) \in S \times I$ such that $r_{s,i} = 1$, we define $C_{s,i} = \sum_{t \in T}r_{s,i,t}$.\\

Latent variables:
\begin{align*}
    f_{i} &= \left\{\begin{tabular}{cl}
        1 & \text{if instance $i$ admits a feasible solution}\\
        0 & \text{otherwise.}
    \end{tabular}
    \right., \quad \forall i \in I\\
    f_{s,i} &= \left\{\begin{tabular}{cl}
        1 & \text{solution $(s,i)$ is feasible}\\
        0 & \text{otherwise.}
    \end{tabular}
    \right., \quad \forall (s,i) \in S \times I \text{ s.t. } r_{s,i} = 1
\end{align*}

Parameters:
% $$\theta = \left(\lambda, (\alpha_s)_{s \in S}, (\beta_s)_{s \in S}, (\gamma_s)_{s \in S}, (\alpha_t)_{t \in T}, (\beta_t)_{t \in T}\right)$$
$$\theta = \left(\lambda, (\alpha_s)_{s \in S}, (\beta_s)_{s \in S}, (\gamma_s)_{s \in S}, a_0, b_0, a_1, b_1\right)$$
where
\begin{alignat*}{3}
    \lambda &= \mathbb{P}\left(f_i = 1\right) && \text{(feasiblity rate of instances)}\\
    \alpha_s &= \mathbb{P}\left(r_{s,i} = 1 \mid f_i = 0\right) && \text{(type I error of solver $s$)}\\
    \beta_s &= \mathbb{P}\left(r_{s,i} = 0 \mid f_i = 1\right) && \text{(type II error of solver $s$)}\\
    \gamma_s &= \mathbb{P}\left(f_{s,i} = 1 \mid f_i = 1, r_{s,i} = 1\right) \quad  && \text{(feasibility rate of solutions reported by $s$ on feasible instances)}\\
    a_0, b_0 & && \text{Parameters of the Beta–Binomial for infeasible solutions}\\
    a_1, b_1 & && \text{Parameters of the Beta–Binomial for feasible solutions}
    % \alpha_t &= \mathbb{P}\left(r_{s,i,t} = 1 \mid f_{s,i} = 0, r_{s,i} = 1\right) && \text{(type I error of test $t$)}\\
    % \beta_t &= \mathbb{P}\left(r_{s,i,t} = 0 \mid f_{s,i} = 1, r_{s,i} = 1\right) \quad\quad && \text{(type II error of test $t$)}\\
\end{alignat*}

\subsection{E-step}
We first compute the intermediate quantities involved in the conditional expectations of $f_i$ and $f_{s,i}$. Let $\textbf{R}$ denote all observed variable and $\textbf{R}_{s} = \{r_{s,i}:\; i \in I\} \cup \left\{ r_{s,i,t} :\; (i,t) \in I \times T, r_{s,i} = 1\right\}$ denote the observed variables related to solver $s$. Let $\nu(k\mid n, a, b) = \binom{n}{k}\frac{B(k+a,n-k+b)}{B(a,b)}$ be the probability mass function of the Beta-Binomial distribution, where $B$ is the beta function. We define for all $(s,i) \in S \times I$:
% \begin{align*}
% A_{s,i}^{(1)} &\triangleq \P{\textbf{R}_s \mid r_{s,i} = 1, f_{s,i} = 1, \th}\\
% &= \prod_{t \in T}(1-\beta_t)^{r_{s,i,t}} \cdot \beta_t^{(1-r_{s,i,t})}
% \end{align*}

% \begin{align*}
% A_{s,i}^{(0)} &\triangleq \P{\textbf{R}_s \mid r_{s,i} = 1, f_{s,i} = 0, \th}\\
% &= \prod_{t \in T}\alpha_t^{r_{s,i,t}} \cdot (1-\alpha_t)^{(1-r_{s,i,t})}
% \end{align*}

\begin{align*}
A_{s,i}^{(0)} &\triangleq \nu (C_{s,i} \mid T, a_0, b_0)\\
A_{s,i}^{(1)} &\triangleq \nu (C_{s,i} \mid T, a_1, b_1)
\end{align*}

and, for all $i \in I$:
\begin{align*}
B_{i}^{(1)} &\triangleq \P{\textbf{R} \mid f_{i} = 1, \th}\\
&= \prod_{s \in S} \left[(1-\beta_s) \left(\gamma_s \P{\textbf{R}_s \mid r_{s,i} = 1, f_{s,i} = 1, \th} + (1-\gamma_s) \P{\textbf{R}_s \mid r_{s,i} = 1, f_{s,i} = 0, \th} \right)\right]^{r_{s,i}} \beta_s^{(1-r_{s,i})}\\
&= \prod_{s \in S} \left[(1-\beta_s) \left(\gamma_s A_{s,i}^{(1)} + (1-\gamma_s) A_{s,i}^{(0)} \right)\right]^{r_{s,i}} \beta_s^{(1-r_{s,i})}\\
\hfill\\
B_{i}^{(0)} &\triangleq \P{\textbf{R} \mid f_{i} = 0, \th}\\
&= \prod_{s \in S} \left[\alpha_s \P{\textbf{R}_s \mid f_{i} = 0, \th} \right]^{r_{s,i}}(1-\alpha_s)^{(1-r_{s,i})}\\
&= \prod_{s \in S} \left[\alpha_s A_{s,i}^{(0)} \right]^{r_{s,i}}(1-\alpha_s)^{(1-r_{s,i})}\\
\end{align*}

We then proceed to compute the conditional expectation of $f_i$ and $f_{s,i}$ for a given $\theta$. For all $i \in I$:
\begin{align*}
    \E{f_i \mid \textbf{R}, \th} &= \P{f_i=1 \mid \textbf{R}, \th}\\
    &= \frac{\P{\textbf{R} \mid f_i=1, \th} \P{f_i=1 \mid \th}}{\P{\textbf{R} \mid \th}}\\
    &= \frac{\P{\textbf{R} \mid f_i=1, \th} \P{f_i=1 \mid \th}}{\P{\textbf{R} \mid f_i=1, \th}\P{f_i=1} + \P{\textbf{R} \mid f_i=0, \th}\P{f_i=0}}\\
    &= \frac{\lambda B_{i}^{(1)}}{\lambda B_{i}^{(1)} + (1-\lambda)B_{i}^{(0)}}
\end{align*}

For all $(s,i) \in S \times I$ such that $r_{s,i} = 1$:
\begin{align*}
    &\P{f_{s,i}=1 \mid f_i = 1, \textbf{R}, \th}\\
    &\quad = \frac{\P{\textbf{R} \mid f_{s,i}=1, f_i = 1, \th} \P{f_{s,i}=1 \mid f_i = 1, \th}}{\P{\textbf{R} \mid f_i = 1, \th}}\\
    &\quad= \frac{\P{\textbf{R} \mid f_{s,i}=1, f_i = 1, \th} \P{f_{s,i}=1 \mid f_i = 1, \th}}{\P{\textbf{R} \mid f_{s,i}=1, f_i = 1, \th}\P{f_{s,i}=1 \mid f_i = 1} + \P{\textbf{R} \mid f_{s,i}=0, f_i = 1, \th}\P{f_i=0 \mid f_i = 1}}\\
    &\quad= \frac{\gamma_s A_{s,i}^{(1)}}{\gamma_s A_{s,i}^{(1)} + (1-\gamma_s)A_{s,i}^{(0)}}\\
\end{align*}
Since $\P{f_{s,i} = 1 \mid f_i = 0} = 0$, we obtain:
\begin{align*}
    \E{f_{s,i} \mid \textbf{R}, \th} &= \P{f_{s,i}=1 \mid f_i = 1, \textbf{R}, \th} \P{f_i = 1 \mid \textbf{R}, \th} + \P{f_{s,i}=1 \mid f_i = 0, \textbf{R}, \th} \P{f_i = 0 \mid \textbf{R}, \th}\\
    &= \frac{ \gamma_s A_{s,i}^{(1)}}{\gamma_s A_{s,i}^{(1)} + (1-\gamma_s)A_{s,i}^{(0)}} \frac{\lambda B_{i}^{(1)}}{\lambda B_{i}^{(1)} + (1-\lambda)B_{i}^{(0)}}\\
\end{align*}

\subsection{M-step}

% $$\P{ \textbf{R}, \textbf{F} \mid \th } = \prod_{i \in I} \P{f_i} \left[\prod_{s \in S} \P{r_{s,i} \mid f_i} \cdot \prod_{\substack{s \in S \\ r_{s,i}=1}} \P{f_{s,i} \mid f_i}  \prod_{t \in T}\P{r_{s,i,t} \mid f_{s,i}}\right]$$

The log-likelihood is given by:
\begin{align*}
    \ln &\; \P{ \textbf{R}, \textbf{F} \mid \th }\\
    &= \sum_{i \in I} \left[ \ln \P{f_i} + \sum_{s \in S} \ln \P{r_{s,i} \mid f_i} + \sum_{\substack{s \in S \\ r_{s,i}=1}} \left[\ln \P{f_{s,i} \mid f_i}  + \sum_{t \in T}\ln \P{r_{s,i,t} \mid f_{s,i}}\right] \right]\\
    &= \sum_{i \in I} \left[ f_i \ln(\lambda) + (1-f_i)\ln(1-\lambda)\right]\\
    & \quad + \sum_{i \in I} \sum_{s \in S} \left[ r_{s,i} \left( f_i \ln(1-\beta_s) + (1-f_i)\ln(\alpha_s)\right) + (1-r_{s,i})\left( f_i \ln(\beta_s) + (1-f_i)\ln(1-\alpha_s) \right) \right]\\
    & \quad + \sum_{i \in I} \sum_{\substack{s \in S \\ r_{s,i}=1}} f_i \left[ f_{s,i} \ln(\gamma_s) + (1-f_{s,i}) \ln(1-\gamma_s) \right]\\
    & \quad + \sum_{i \in I} \sum_{\substack{s \in S \\ r_{s,i}=1}} \left[ f_{s,i} \ln \nu(C_{s,i}\mid T, a_1, b_1) + (1-f_{s,i}) \ln \nu(C_{s,i}\mid T, a_0, b_0)\right]
    % & \quad + \sum_{i \in I} \sum_{\substack{s \in S \\ r_{s,i}=1}} \sum_{t \in T} \left[ r_{s,i,t} \left( f_{s,i} \ln (1-\beta_t) + (1-f_{s,i})\ln(\alpha_t) \right) + (1-r_{s,i,t}) \left( f_{s,i} \ln(\beta_t) + (1-f_{s,i}) \ln(1-\alpha_t) \right) \right]
\end{align*}

Let $\hat{f}_i = \E{f_i \mid \textbf{R}, \theta}$ and $\hat{f}_{s,i} = \E{f_{s,i} \mid \textbf{R}, \theta} = \E{f_i f_{s,i} \mid \textbf{R}, \theta}$.
Taking the conditional expectation and differentiating with respect to $\lambda$, we have:
\begin{equation}\label{eq:deriv_lambda}
    \frac{\partial}{\partial \lambda} \E{\ln \P{ \textbf{R} \mid \th }} = \sum_{i \in I} \left(\frac{\hat{f}_i}{\lambda} - \frac{1-\hat{f}_i}{1-\lambda}\right)
\end{equation}
Equating \eqref{eq:deriv_lambda} to 0, we obtain: $\lambda = \frac{\sum_{i \in I} \hat{f}_i}{I|}$.
With analogous reasoning, we have:
\begin{align*}
    \alpha_s &= \frac{\sum_{i \in I} (1-\hat{f}_i) r_{s,i}}{\sum_{i \in I} (1-\hat{f}_i)}\\
    \beta_s &= \frac{\sum_{i \in I} \hat{f}_i (1-r_{s,i})}{\sum_{i \in I} \hat{f}_i}\\
    \gamma_s &= \frac{\sum_{i \in I} r_{s,i} \hat{f}_{s,i}}{\sum_{i \in I} r_{s,i }\hat{f}_i}
    % \alpha_t &= \frac{\sum_{i \in I} \sum_{s \in S} r_{s,i} (1-\hat{f}_{s,i}) r_{s,i,t}}{\sum_{i \in I} \sum_{s \in S} r_{s,i} (1-\hat{f}_{s,i})}\\
    % \beta_t &= \frac{\sum_{i \in I} \sum_{s \in S} r_{s,i} \hat{f}_{s,i} (1-r_{s,i,t})}{\sum_{i \in I} \sum_{s \in S} r_{s,i} \hat{f}_{s,i}}
\end{align*}

We use the method of moments to update the parameters $a_0$, $b_0$, $a_1$, and $b_1$ using a Beta prior with $(\bar{\alpha}, \bar{\beta}) = (20,1)$ for the true positive rate. Define:
\begin{align*}
    p_0 &= \frac{\sum_{s,i} \left(1-\hat{f}_{s,i}\right) \frac{C_{s,i}}{T}}{\sum_{s,i} \left(1-\hat{f}_{s,i}\right)}\\
    p_1 &= \frac{\sum_{s,i} \hat{f}_{s,i} \frac{C_{s,i}}{T} + \bar{\alpha} - 1}{\sum_{s,i} \hat{f}_{s,i} + \bar{\alpha} + \bar{\beta} - 2}
\end{align*}
and
\begin{align*}
    \mu_0 &= \frac{\sum_{s,i} \left(1-\hat{f}_{s,i}\right) C_{s,i}}{T \cdot \sum_{s,i} \left(1-\hat{f}_{s,i}\right)}\\
    \sigma^2_0 &= \frac{\sum_{s,i} \left(1-\hat{f}_{s,i}\right) \left(C_{s,i}/T - \mu_0\right)^2}{\sum_{s,i} \left(1-\hat{f}_{s,i}\right)}\\
    \mu_1 &= \frac{\sum_{s,i} \hat{f}_{s,i} C_{s,i}}{T \cdot \sum_{s,i} \hat{f}_{s,i}}\\
    \sigma^2_1 &= \frac{\sum_{s,i} \hat{f}_{s,i} \left(C_{s,i}/T - \mu_1\right)^2}{\sum_{s,i} \hat{f}_{s,i}}
\end{align*}

Then for $k \in \{0,1\}$:
\begin{align*}
    \rho_k = \frac{\frac{T \sigma_k^2}{\mu_k(1-\mu_k)}-1}{T-1}
\end{align*}
Finally, retrieve $a_k$ and $b_k$ from $p_k$ and $\rho_k$ for $k \in \{0,1\}$ as:
\begin{align*}
    a_k &= p_k \left(\frac{1}{\rho_k} - 1\right)\\
    b_k &= (1-p_k) \left(\frac{1}{\rho_k} - 1\right)\\
\end{align*}

\end{document}